%% file: main.tex
\documentclass[runningheads]{llncs}
\usepackage[T1]{fontenc}
\usepackage{graphicx}
\usepackage{booktabs}
\usepackage[misc]{ifsym}
\newcommand{\corr}{(\Letter)}

\usepackage{algorithm,algorithmic, amsmath, amsfonts}
\usepackage{multirow, booktabs, graphicx, subcaption}
\usepackage{color}
\usepackage{pdflscape, rotating}
\usepackage{pgfplotstable}
\usepackage{booktabs}
\usepackage{placeins}
\usepackage{makecell}

\begin{document}

\title{Efficient Zero-Shot AI-Generated Image Detection}


\author{ Ryosuke Sonoda\inst{1} \corr \and
 Ramya Srinivasan\inst{2}}

\authorrunning{R. Sonoda and R. Srinivasan}

\institute{Fujitsu Ltd., Japan \email{sonoda.ryosuke@fujitsu.com}
\and
Fujitsu Research of America, Inc., USA \email{ramya@fujitsu.com}}

\maketitle              

\input{sections/abstract}    
\input{sections/introduction}
\input{sections/related_work}
\input{sections/method}
\input{sections/experiment}
\input{sections/conclusion}



\bibliographystyle{splncs04}
\bibliography{mybibliography}

\end{document}

%% file: sections/abstract.tex
\begin{abstract}
The rapid progress of text-to-image models has made AI-generated images increasingly realistic, posing significant challenges for accurate detection of generated content. 
While training-based detectors often suffer from limited generalization to unseen images, training-free approaches offer better robustness, yet struggle to capture subtle discrepancies between real and synthetic images.
In this work, we propose a training-free AI-generated image detection method that measures representation sensitivity to structured frequency perturbations, enabling detection of minute manipulations. 
The proposed method is computationally lightweight, as perturbation generation requires only a single Fourier transform for an input image.
As a result, it achieves one to two orders of magnitude faster inference than most training-free detectors.
Extensive experiments on challenging benchmarks demonstrate the efficacy of our method over state-of-the-art (SoTA).
In particular, on OpenFake benchmark, our method improves AUC by nearly $10\%$ compared to SoTA, while maintaining substantially lower computational cost.
\keywords{Fake image detection  \and Zero-shot \and Generative AI.}
\end{abstract}

%% file: sections/introduction.tex
\section{Introduction}
\label{sec:intro}
Rapid advancements in generative AI has brought with it both opportunities and challenges. While generative AI has ushered productivity gains across multiple sectors \cite{park_2023}, its proliferation has also increased the chances of various types of adversarial attacks and vulnerabilities \cite{kokil}. The rise of AI generated deepfakes \cite{yisroel}, cyber espionage \cite{anthropic}, computer network attacks \cite{baiqiang}, sockpuppet accounts \cite{kyang}, and comment brigading \cite{jiawei} are just a handful of the many ways in which generative AI is being misused by malicious players. 

As a consequence, a variety of counter-measures are being developed towards detecting, mitigating, and preventing mis-use of generative AI. These include development of novel defense mechanisms, regulatory and policy measures \cite{scientific,nadia}, among others.  In particular, with the emergence of text to image (T2I) and text to video (T2V) foundation models, creating fake content has become easier than ever before. Thus, a major area of focus has been in detecting AI generated content (audio, text, images, and videos) \cite{cui,yu,alam,kaiqing,ziqiang}. 

Our work complements and augments aforementioned efforts in detecting AI generated content. In particular, the proposed method detects AI generated images in a {\it training-free manner} across a wide range of generation processes (SoTA diffusion and GAN based methods) and image domains (e.g., politics, society news), thereby demonstrating the generalizability and robustness of detection performance and applicability. Furthermore, the proposed method is computationally lightweight, achieving upto two orders of magnitude faster inference time when compared to most training-free detectors, thus making it a viable choice for deployment on edge devices. 

\begin{figure}[h]
    \centering

    \begin{minipage}[t]{0.7\linewidth}
        \centering
        \includegraphics[width=\linewidth]{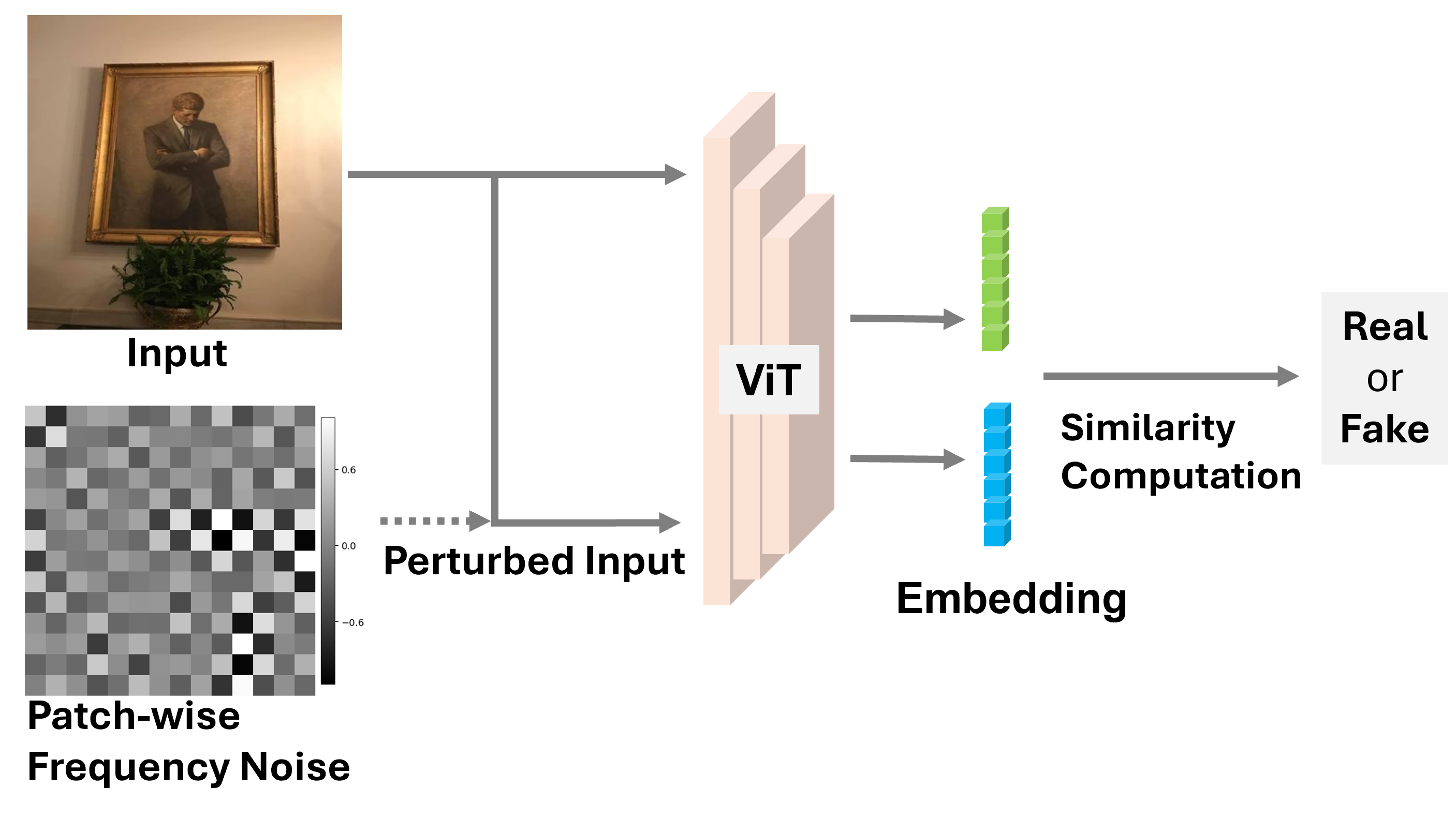}
        \subcaption{Overall procedure of our method.}
    \end{minipage}
    \hfill
    \begin{minipage}[t]{0.28\linewidth}
        \centering
        \includegraphics[width=\linewidth]{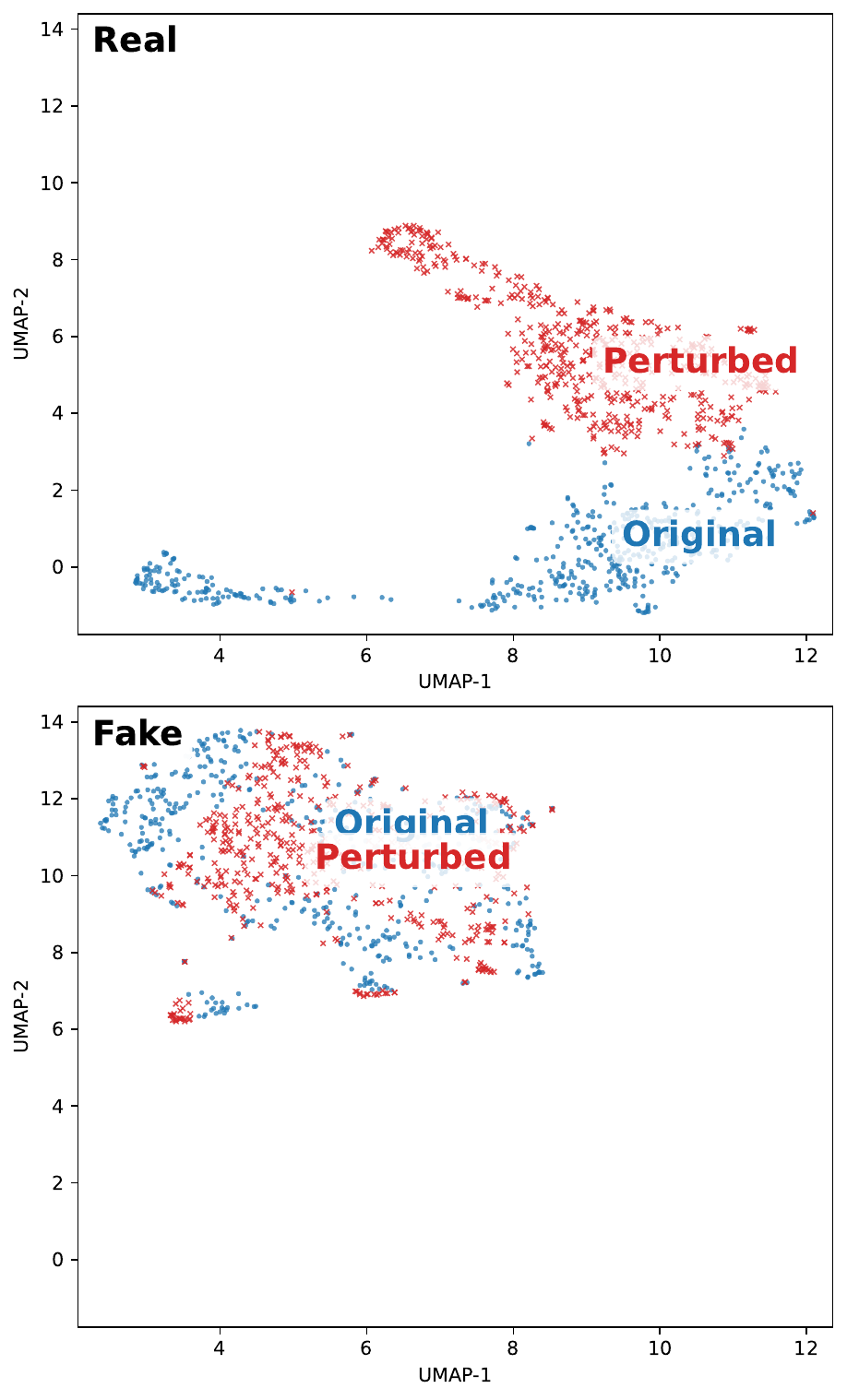}
        \subcaption{UMAP visualization.}
    \end{minipage}
    \caption{(a) Our method first generates a perturbed input by adding frequency-band-limited noise to each patch of the input image. These inputs are fed to Vision Transformer (ViT), then a Real/Fake classification is performed based on the distance between feature representations.
    (b) UMAP \cite{umap} projection of CLIP embeddings for original (blue, circle) and perturbed (orange, cross) images. 
    The top panel corresponds to real images, and the bottom corresponds to fake (generated) images. As can be observed, real data exhibits distinct regions of original and perturbed images as opposed to fake data. For visualization purposes, we randomly sub-sampled 1,000 samples from OpenFake dataset}
    \label{fig:method}
\end{figure}

Our method is motivated by the observation that synthetic images contain frequency artifacts that cause systematically different responses in Vision Foundation Model (VFM) representations compared to real images. 
We therefore probe images using structured frequency perturbations and quantify the resulting representation sensitivity to distinguish real images from synthetic ones (Figure~\ref{fig:method}). 
Unlike prior training-free detectors that rely on large numbers of perturbed samples or repeated generative reconstruction, our method requires only a single Fourier transform and a single forward pass of an ViT, resulting in substantially lower computational cost. 
Extensive experiments on OpenFake~\cite{openfake}, GenImage~\cite{genimage}, 
and Semi-Truth~\cite{semitruth} benchmarks spanning over a dozen of generative models demonstrate the improvement in detection accuracy and generalizability of the proposed approach over SoTA training-free detection methods~\cite{dtad,mib,aeroblade,warpad,rigid,minder,zerofake}.



%% file: sections/related_work.tex
\section{Related Work}
\label{sec:related}

Detecting images generated by text-to-image (T2I) models has become an increasingly critical problem with the rapid advancement of generative models. 
Existing approaches for AI-generated image detection can be broadly divided into training-based and training-free methods.

Training-based approaches rely on supervised learning with large-scale datasets containing both real and synthetic images, aiming to capture artifacts or biases introduced during the generation process. 
A line of work focuses on frequency-based cues, exploiting artificial features left by generators such as those generated by up-sampling in the frequency domain ~\cite{npr,chu2025fire}. 
Other methods utilize reconstruction errors, where models are trained to reconstruct images, and detection is performed based on reconstruction discrepancies between real and generated images~\cite{wang2023dire}. 
With the emergence of T2I models, recent studies further explore vision--language relationship by leveraging vision-language models such as CLIP to detect inconsistencies between images and their associated text prompts~\cite{c2pclip,defake,fakeinversion}. 
Despite their effectiveness within the training domain, these methods often struggle to generalize to unseen generative models. 

In contrast, training-free methods aim to achieve zero-shot generalization without relying on labeled training data. 
Reconstruction-based approaches employ pre-trained autoencoders or diffusion models to measure reconstruction discrepancies, under the assumption that real images deviate from the manifold of generators~\cite{aeroblade,dtad,mib}. 
Also, adversarial text-driven reconstruction methods incorporate T2I generation priors to enable training-free detection~\cite{zerofake}. 
Another line of work focuses on frequency-based analysis, where high-frequency components are used to identify statistical anomalies in generated images~\cite{hfi,reper}. 
Perturbation-based methods further assess the robustness of image representations under controlled perturbations, often leveraging large pre-trained vision models to expose representation instability in real images~\cite{rigid,minder}.

Our approach follows the training-free paradigm but differs from existing methods by considering structured frequency-specific perturbations and semantic representations from a vision-language foundation model. 
By considering the sensitivity of the model to structured high-frequency perturbations, our method captures fine-grained discrepancies at a much lower computational cost compared to most SoTA training-free methods.

%% file: sections/method.tex
\section{Method}
\label{sec:method}
We aim to distinguish real images from synthetic ones by probing their response to structured high-frequency perturbations.
In a training-free setting
, we define a score function that quantifies the sensitivity of VFM representations to controlled perturbations in the frequency domain (Section~\ref{subsec:score}).
We adopt the vision encoder of CLIP~\cite{clip} as a feature extractor and keep its parameters fixed during inference, while the choice of representation layer is detailed in Section~\ref{subsec:layer}.




\subsection{AI-generated Image Detection}
\label{subsec:score}

We propose to detect AI-generated images by analyzing their sensitivity of visual representations to structured high-frequency perturbations. 
Our underlying hypothesis is that compared to real images, synthetic images exhibit characteristic frequency biases induced by generator architectures and training data distribution, which may lead to different responses under targeted frequency perturbations.

Given an input image $x \in \mathbb{R}^{H \times W \times C}$, we construct a frequency perturbation $\delta$ restricted to the high-frequency band in the Fourier domain.
Specifically, we sample Gaussian noise $\epsilon \sim \mathcal{N}(0, \lambda I)$ where $\lambda > 0$ controls the perturbation strength in the pixel space, transform it using the fast Fourier transform (FFT), suppress its low-frequency components, and then invert it back to obtain a high-frequency perturbation $\delta$.
The perturbed image is defined as
\begin{equation}
\tilde{x} = x + \delta.
\label{eq:fre_noise}
\end{equation}
We define high-frequency components as the Fourier coefficients whose radial frequency exceeds a threshold $\tau$ in the normalized frequency domain.
Unlike real images, whose high-frequency components encode diverse details, generated images often inherit frequency biases induced by generator priors and training objectives.
Consequently, perturbations along high-frequency directions are expected to alter real images more substantially.

To control the spatial scale of perturbations, we optionally partition the image into non-overlapping patches of size $P \times P$ and apply the above frequency perturbation independently within each patch.
Let $x \in \mathbb{R}^{H \times W \times C}$ be partitioned into $N$ non-overlapping patches $\{x_i\}_{i=1}^{N}$ of spatial size $P \times P$, where $N = \frac{H}{P} \times \frac{W}{P}$.
For each patch $x_i$, we independently construct a high-frequency perturbation $\delta_i$
using the procedure described above, and obtain the perturbed patch $\tilde{x}_i = x_i + \delta_i$.
The final perturbed image $\tilde{x}$ is reconstructed by assembling all $\tilde{x}_i$.
This formulation allows the spatial correlation length of perturbations to be controlled via the patch size $P$.

Let $f(\cdot)$ denote the embedding function of a fixed pre-trained VFM.
We define the detection score $S(x)$ as the cosine similarity between the embeddings of the original and perturbed images:
\begin{equation}
S(x) = \mathrm{SIM}\bigl(f(x), f(\tilde{x})\bigr),
\label{eq:score}
\end{equation}
where $\mathrm{SIM}(\cdot, \cdot)$ is the cosine similarity.

A higher score indicates that the image representation is invariant to structured high-frequency perturbations. 
Based on our hypothesis, AI-generated images tend to yield higher similarity scores due to their constrained high-frequency characteristics, whereas real images exhibit larger representation shifts.

\subsection{Layer-wise Representation Choice}\label{subsec:layer}
A common approach for image representation in CLIP~\cite{clip} is to use the final layer's output embedding, which is highly effective for semantic alignment but often suboptimal for detecting subtle visual discrepancies.
CLIP employs a hierarchical structure: shallow layers capture low-level features (edges, textures), while deeper layers progressively abstract to high-level semantic concepts. 
The final layer's CLS token is explicitly optimized to be invariant to fine-grained appearance variations that do not alter the core semantic meaning.

However, our detection framework relies on structured high-frequency perturbations, which predominantly affect fine-scale image details while largely preserving global semantics. 
Therefore, representations that overly emphasize semantic invariance may suppress the signal induced by our perturbation strategy.

To address this mismatch, we extract image representations from an intermediate layer $l$ of CLIP. 
Compared to the final embedding, intermediate-layer features retain finer-grained visual cues while preserving sufficient semantic structure, making them better suited for capturing representation shifts induced by high-frequency perturbations. 
Unless otherwise stated, we therefore use intermediate-layer CLIP embeddings for computing the similarity score in Eq.~\eqref{eq:score}.
We also discuss performance differences across each layer in our ablation study.

\subsection{Computational Complexity}\label{subsec;computation}
Several training-free detection methods follow a common paradigm: 
they generate perturbed versions of the input image, feed both the original and perturbed images into a VFM, and compute discrepancies in the VFM feature space. 
Accordingly, their computational cost is primarily determined by (i) the cost of generating perturbed samples and (ii) the cost of VFM forward passes.

Let $C_{\text{VFM}}$ denote the cost of a single VFM forward pass.
In our method, perturbation generation requires one 2D FFT and inverse FFT, resulting in $\mathcal{O}(HW \log(HW))$ complexity, followed by a single backbone inference.
The overall complexity is therefore $\mathcal{O}(HW \log(HW)) + C_{\text{VFM}}$.
In practice, the cost is dominated by a single backbone forward pass.

In contrast, methods such as WARPAD~\cite{warpad} evaluate $K$ perturbed samples per image, requiring $K$ forward passes per input. 
Approaches requiring reconstruction of the input image~\cite{aeroblade,zerofake,mib,hfi,dtad} necessitate multiple invocations of generative models, further increasing computational overhead. 
These structural differences account for the runtime gap reported in Section~\ref{subsec: main_result}.





%% file: sections/experiment.tex
\section{Experiments}
We first introduce the datasets, generative models and the baseline methods considered in our analysis (Section~\ref{subsec: exp_set}).
We report the performance of our method across these datasets (Section~\ref{subsec: main_result}). 
Finally, we present detailed ablation studies of our method and test its robustness across data corruptions (Section~\ref{subsec:ablation}).
\subsection{Experimental Settings}\label{subsec: exp_set}

\noindent\textbf{Datasets:}
We evaluate our method on several widely used datasets for AI-generated image detection, covering diverse generative models, image domains, and evaluation settings.
Specifically, we conduct experiments on OpenFake~\footnote{https://huggingface.co/datasets/ComplexDataLab/OpenFake}~\cite{openfake}, GenImage~\footnote{https://github.com/GenImage-Dataset/GenImage}~\cite{genimage}, 
and Semi-Truth~\footnote{https://huggingface.co/datasets/semi-truths/Semi-Truths-Evalset}~\cite{semitruth} which provide large-scale test sets generated by a wide range of diffusion- and GAN-based models, enabling comprehensive evaluation across heterogeneous generation processes. 
We adopt the official test split provided by the dataset, and no training data is used, as all compared methods are entirely training-free.
For Semi-Truth dataset, samples with missing labels or corrupted image files were excluded to ensure evaluation integrity.
Table~\ref{tab:dataset} summarizes the key statistics of the datasets used in our experiments, including the number of generative models and test images.

The three datasets differ substantially in the number of generative models employed.
Semi-Truth contains images generated by a limited set of diffusion-based models, including Kandinsky 2.2, OpenJourney, and Stable Diffusion variants.
GenImage further expands generator diversity, including diffusion and GAN-based architectures such as ADM, BigGAN, Glide, VQDM, Wukong, Midjourney, and Stable Diffusion, resulting in substantial heterogeneity across generative paradigms.
OpenFake includes images synthesized by a large number of generators, spanning both commercial and open-source systems, making it the most diverse and challenging dataset in terms of generator coverage.
This variation in generator composition enables evaluation under heterogeneous generative distributions and differing levels of model diversity.

\begin{table}[t]
    \centering
    \caption{Statistics of datasets in our experiments.}
    \begin{tabular}{cccc}\hline
        Dataset &  $\#$ of Generators & $\#$ of Real& $\#$ of Generated  \\ \hline \hline
        Openfake & 34 & 29829 & 29829 \\
        GenImage & 8 & 50000 & 50000 \\
        Semi-Truth & 5 & 6967 & 26786 \\ \hline
    \end{tabular}
    \label{tab:dataset}
\end{table}

\noindent\textbf{Evaluation Metrics:}
Following prior work on training-free AI-generated image detection, we report detection performance using the Area Under the Receiver Operating Characteristic curve (AUC), which is threshold-independent.
We also report on the runtime to analyze the computational efficacy of various methods.

\noindent\textbf{Baselines:}
We compare our method with a broad set of SoTA training-free detectors.
These baselines span multiple methodological categories, including reconstruction-based methods (AEROBLADE~\cite{aeroblade}, DTAD~\cite{dtad}, MIBD~\cite{mib}, ZEROFAKE~\cite{zerofake}), frequency-based detectors (HFI~\cite{hfi}, WARPAD~\cite{warpad}), perturbation-based methods (RIGID~\cite{rigid}, MINDER~\cite{minder}).
All baseline results are obtained using official implementations or reported settings to ensure fair comparison.

\noindent\textbf{Implementation Details:}
In our method, all input images are resized to $224 \times 224$ as we use CLIP ViT-L/14~\footnote{https://huggingface.co/openai/clip-vit-large-patch14} as the vision foundation model.
For localized frequency perturbation, we divide each image into $N=16$ non-overlapping patches with patch-size $P=14$.
We extract CLS embeddings from an intermediate transformer layer $l=13$ of CLIP ViT.
The high-frequency threshold is set to $\tau=0.5$.
All methods are implemented in PyTorch, and experiments are conducted on a single NVIDIA A100 GPU.
The batch-size is fixed at $8$ for all experiments, and the random seed is also fixed for reproducibility.

\subsection{Main Results}\label{subsec: main_result}
\begin{table}[ht]
    \centering
    \caption{AUC scores of all methods evaluated on the OpenFake dataset.
    Rows correspond to image generators, columns to detection methods, and the final row reports the average AUC across generators. The best results are highlighted in bold.}
    \label{tab:openfake}

\input{tables/openfake}
\end{table}


\begin{table*}[ht]
    \centering
    \caption{AUC scores of all methods evaluated on the Semi-Truth dataset.}
    \label{tab:semitruth}
    \input{tables/semitruth}
\end{table*}

\begin{table*}[ht]
    \centering
    \caption{AUC scores of all methods evaluated on the Genimage dataset.}
    \label{tab:genimage}
    \input{tables/genimage}
\end{table*}

\begin{figure}
    \centering
    \begin{minipage}{0.32\linewidth}
        \centering
        \includegraphics[width=\linewidth]{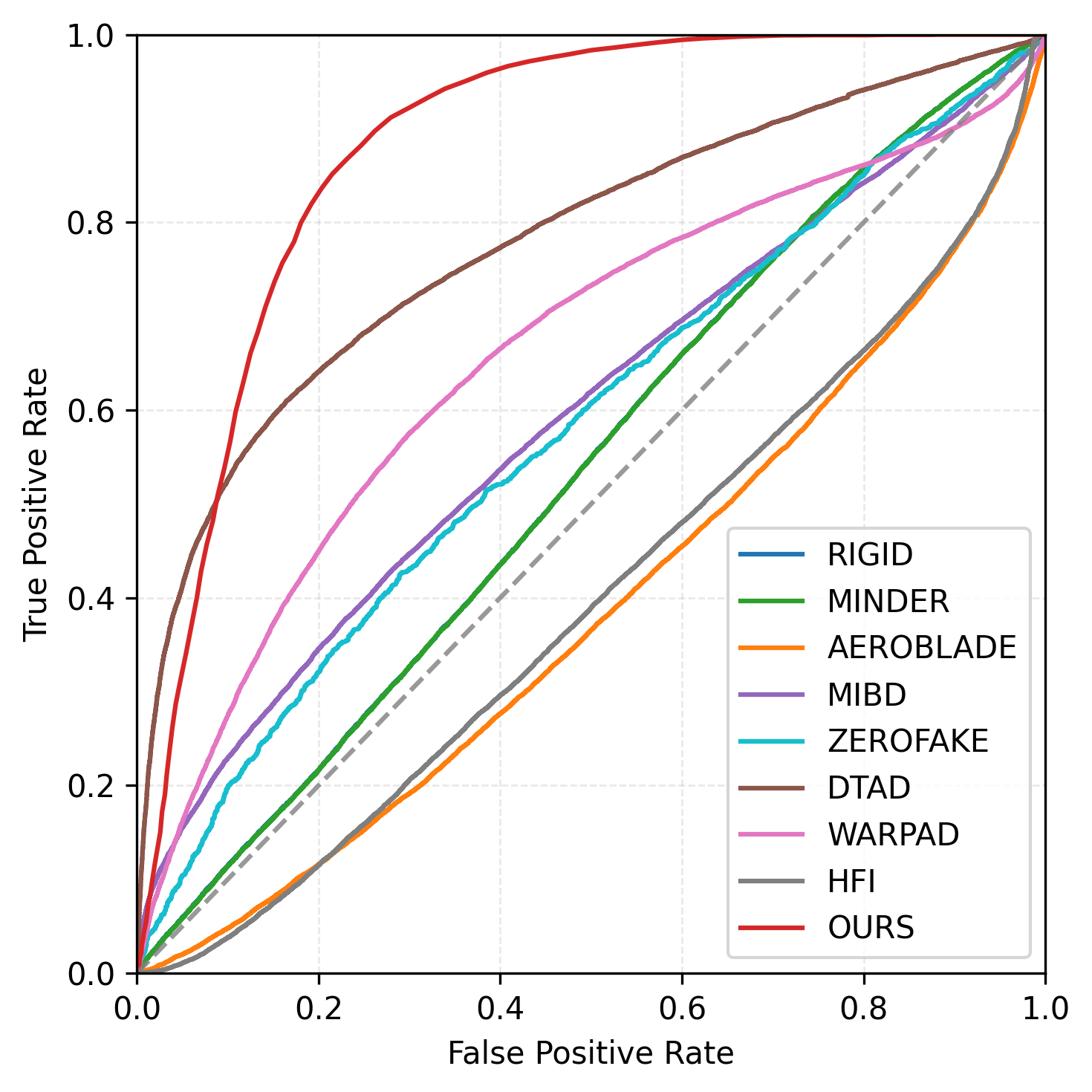}
        \subcaption{Openfake dataset.}
    \end{minipage}
    \begin{minipage}{0.32\linewidth}
        \centering
        \includegraphics[width=\linewidth]{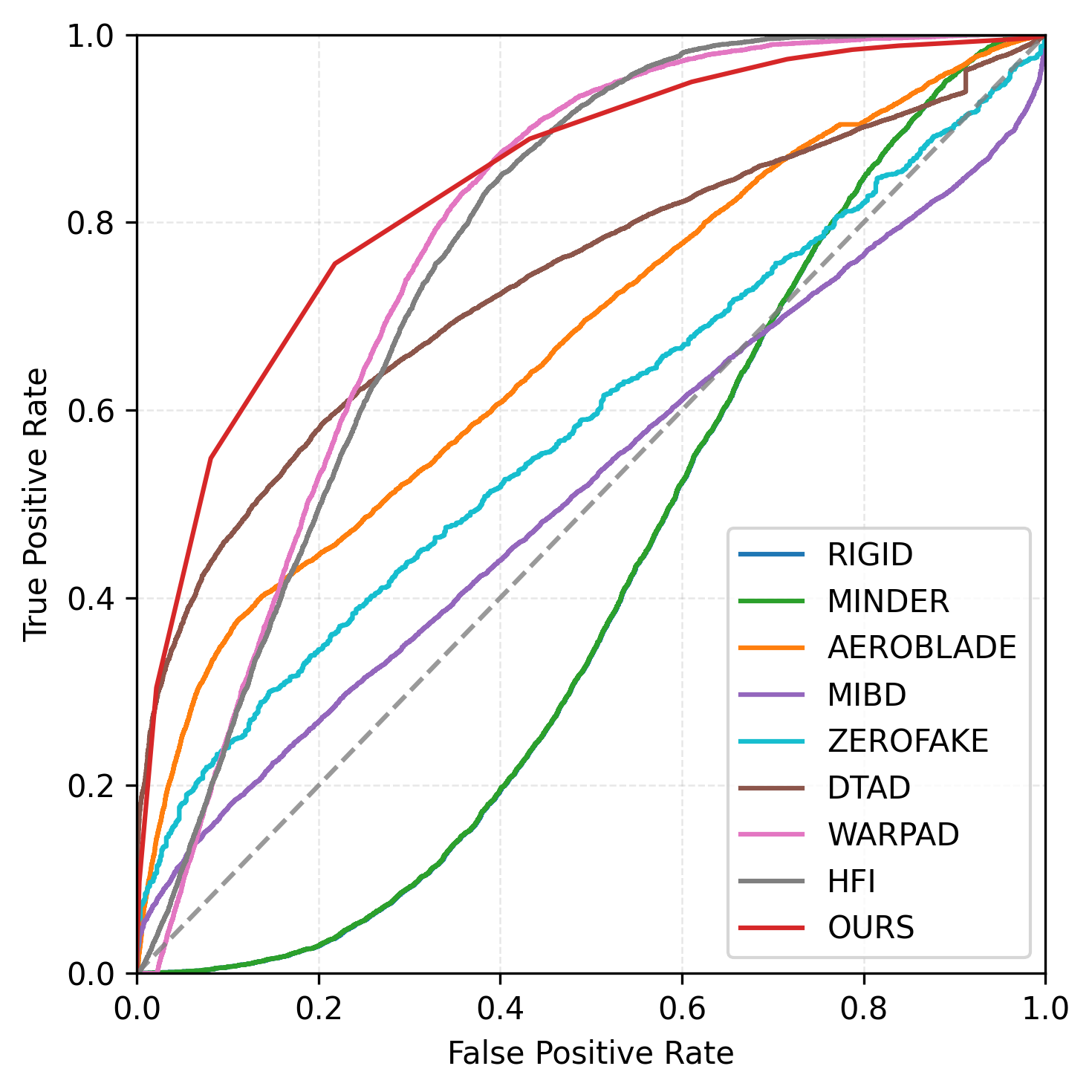}
        \subcaption{Semi-Truth dataset.}
    \end{minipage}
    \begin{minipage}{0.32\linewidth}
        \centering
        \includegraphics[width=\linewidth]{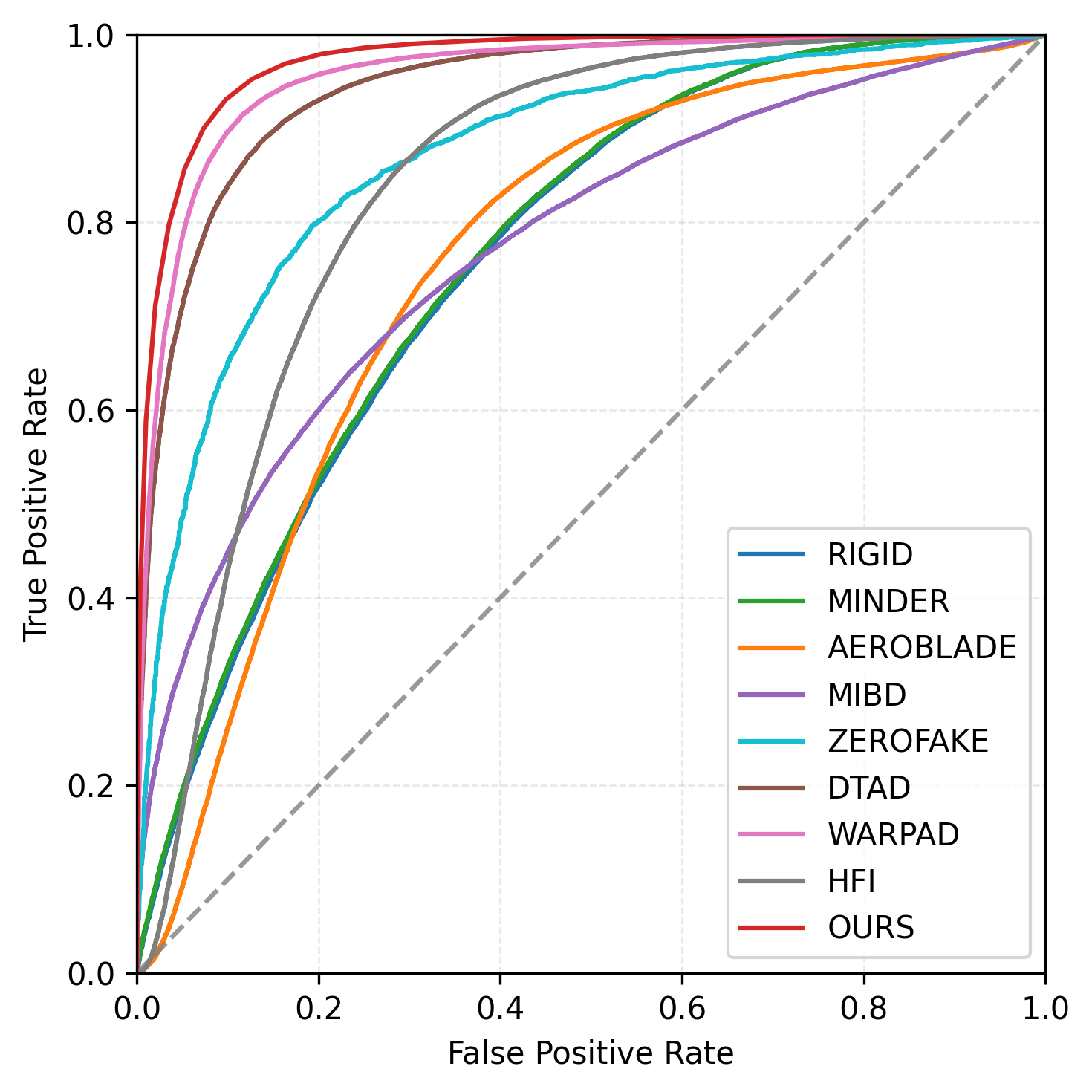}
        \subcaption{Genimage dataset.}
    \end{minipage}
    \caption{ROC curves for the Openfake, Semi-Truth, and Genimage dataset.}
    \label{fig:curves}
\end{figure}

Tables~\ref{tab:openfake}
, \ref{tab:semitruth}, and \ref{tab:genimage}
summarize the performance of our method against existing training-free approaches
on the OpenFake, 
Semi-Truth, and GenImage datasets.
Across all three datasets, our method consistently achieves the highest average AUC.
In particular, we observe an average AUC improvement of $10\%$ over the DTAD on openfake, with gains of up to $14\%$ on Semi-Truth.
Unlike competing methods, whose performance varies substantially across datasets, our approach maintains stable performance under diverse generative settings.
These results indicate that our method generalizes robustly across datasets with different generation sources and distributional characteristics.
We additionally report ROC curves on the Openfake, Semi-Truth, and GenImage datasets in Figure~\ref{fig:curves}. 
The ROC curves show that our method consistently outperforms existing approaches across a broad range of false-positive rates, indicating robustness against distributional shifts. 

\begin{figure}[ht]
    \centering
    \begin{minipage}[gt]{0.32\linewidth}
        \centering
        \includegraphics[width=\linewidth]{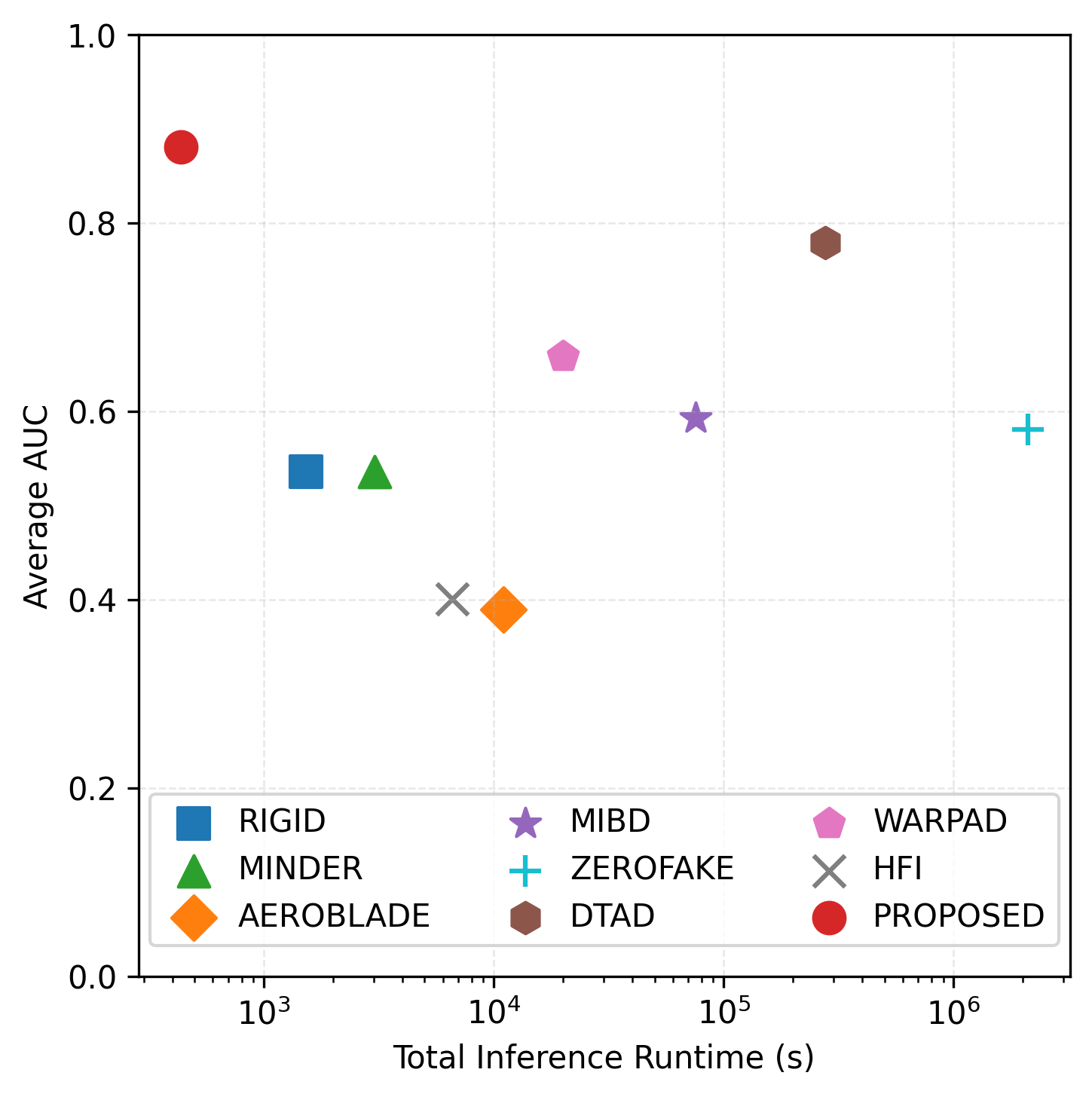}
        \subcaption{Openfake dataset.}
    \end{minipage}
    \hfill
    \begin{minipage}[ht]{0.32\linewidth}
        \centering
        \includegraphics[width=\linewidth]{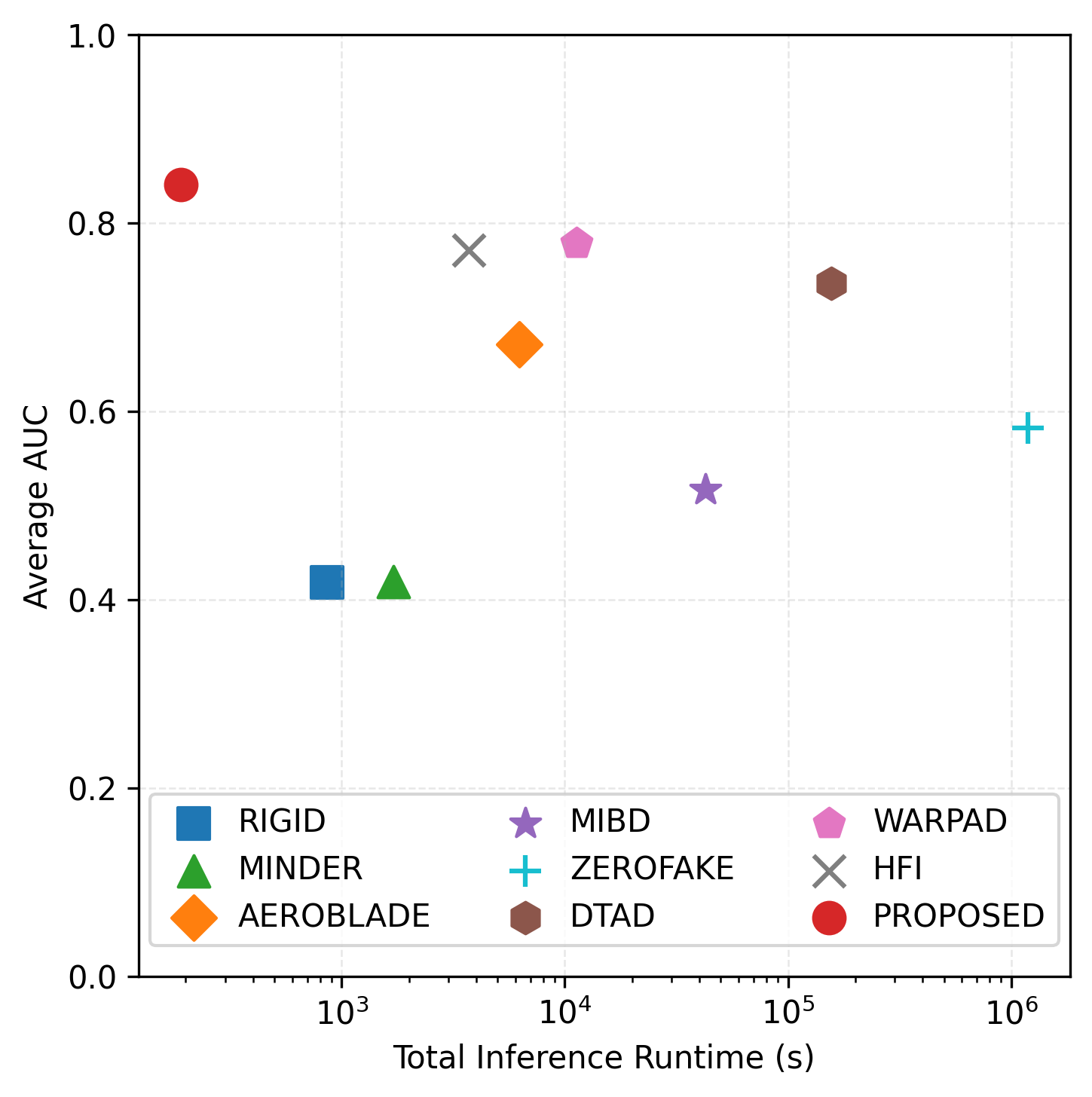}
        \subcaption{Semi-Truth dataset.}
    \end{minipage}
    \hfill
    \begin{minipage}[gt]{0.32\linewidth}
        \centering
        \includegraphics[width=\linewidth]{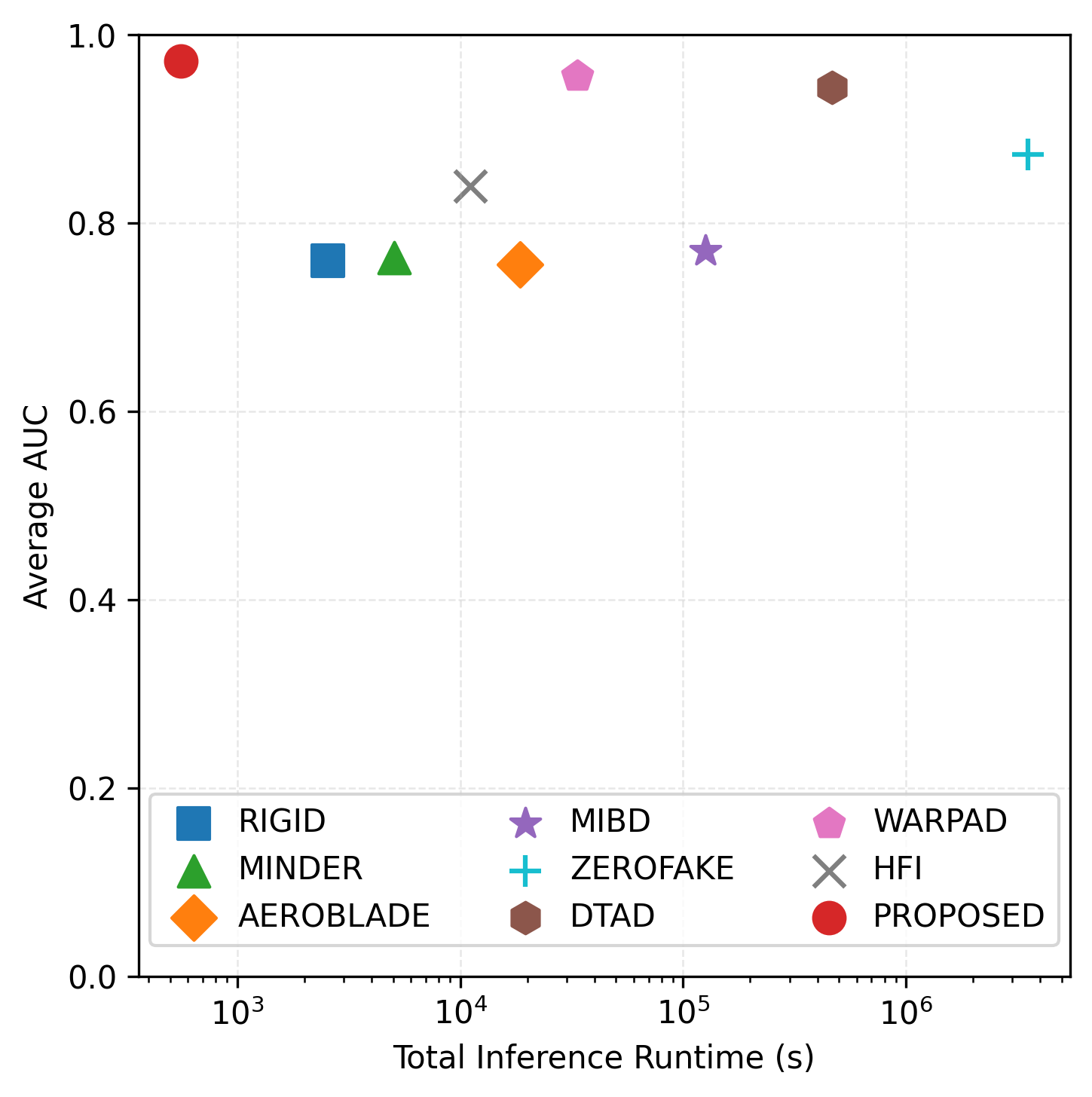}
        \subcaption{Genimage dataset.}
    \end{minipage}

    \caption{Average AUC versus total inference runtime (batch size = 8).
    Higher AUC and lower runtime are preferred; points closer to the upper-left corner indicate better accuracy-speed performance.}
    \label{fig:runtime}
\end{figure}

Figure~\ref{fig:runtime} shows the average AUC scores versus inference runtime of each method on three datasets, measured on a single NVIDIA A100 GPU.
The corresponding numerical values of runtime are provided in Table~\ref{tab:runtime} 
for completeness and reproducibility.
Runtime is computed from model input to prediction output, excluding image preprocessing with a mini-batch size of 8.
Our method achieves the lowest computational cost among all compared approaches, providing one to two orders of magnitude faster inference than prior SoTA methods.
Compared to RIGID—the second fastest method—our approach is over $2\times$ faster while simultaneously achieving substantially higher detection performance, with an AUC improvement of approximately $40\%$ on OpenFake.
The runtime gap between RIGID and our method appears to stem primarily from differences in backbone inference speed, as RIGID employs a relatively simple perturbation design.
Compared to other training-free baselines, our method remains significantly more efficient.
Although WARPAD and DTAD attain competitive AUC scores, their computational overhead is significantly higher, being tens to hundreds of times slower than our method.
This inefficiency stems from large-scale perturbed-sample evaluation in WARPAD and iterative noise sampling with repeated denoising steps in DTAD.

Overall, these results demonstrate that our method achieves competitive detection performance while maintaining computational efficiency.

\begin{table}[t]
    \centering
    \caption{Total inference runtime (in seconds) with a batch size of 8.}
    \label{tab:runtime}
    \input{tables/runtime}
\end{table}

\FloatBarrier

\begin{figure}[ht]
    \centering
    \begin{minipage}[gt]{0.48\linewidth}
        \centering
        \includegraphics[width=\linewidth]{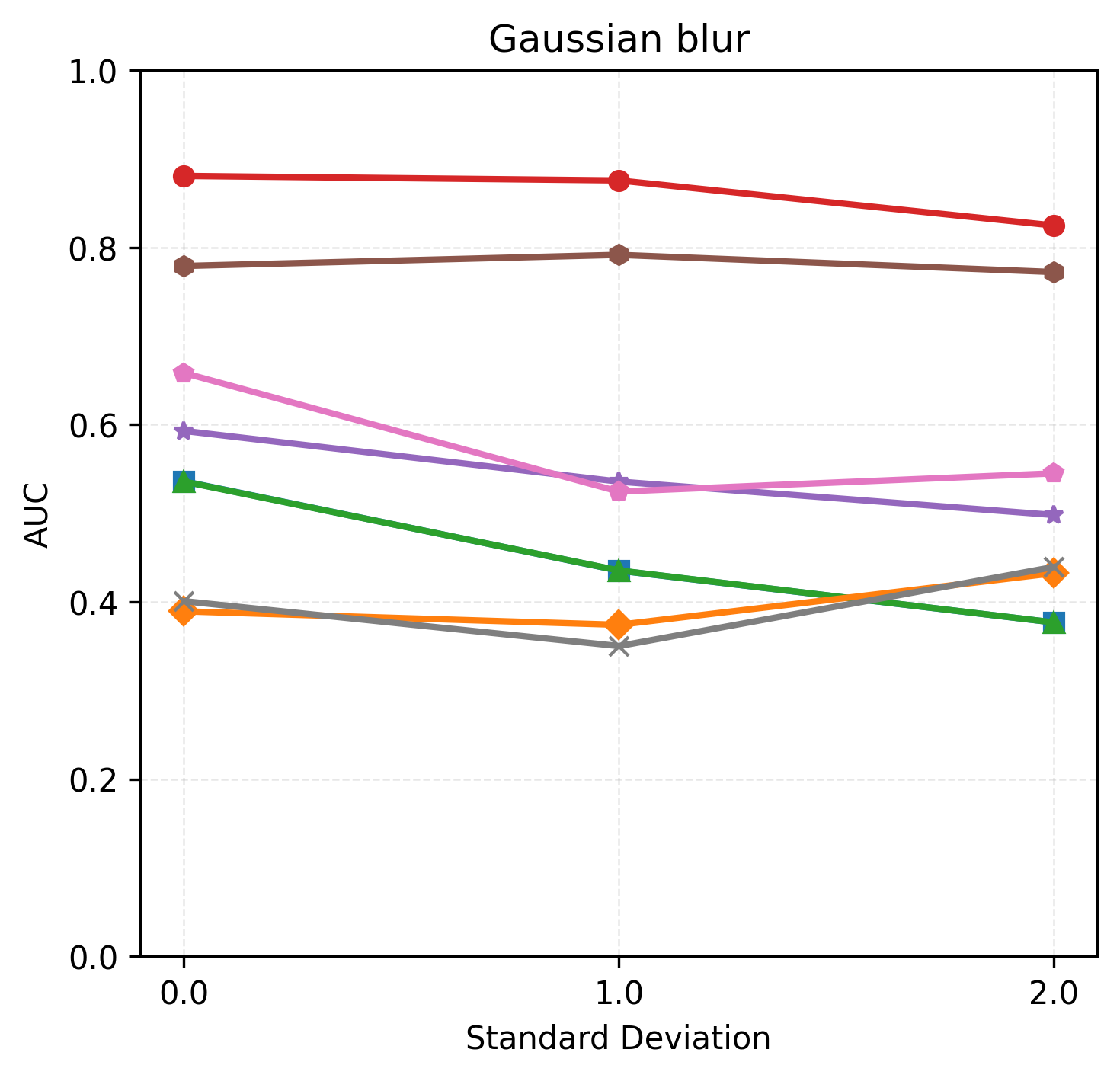}
        \subcaption{Gaussian Blur}
    \end{minipage}
    \hfill
    \begin{minipage}[ht]{0.48\linewidth}
        \centering
        \includegraphics[width=\linewidth]{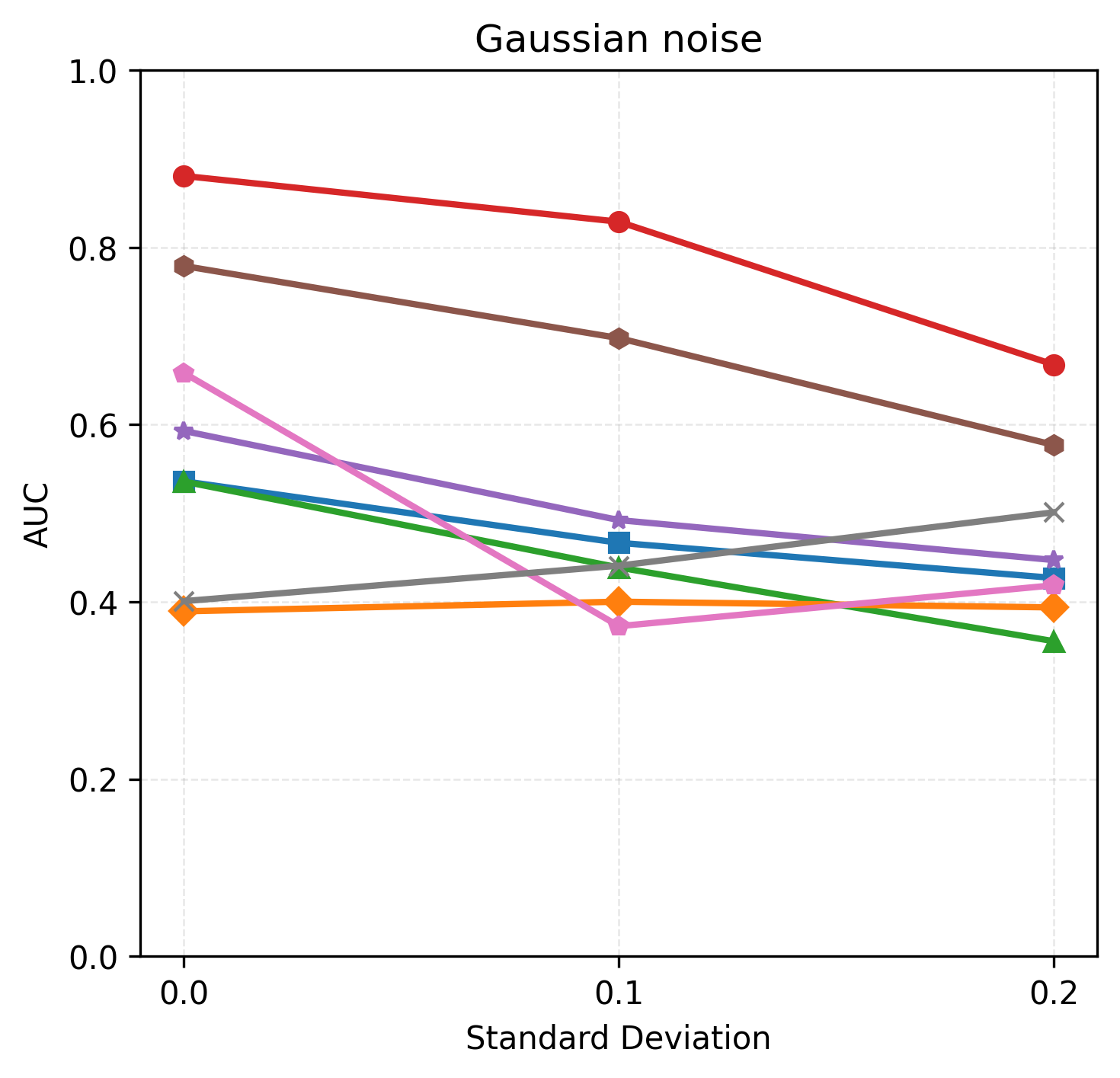}
        \subcaption{Gaussian Noise}
    \end{minipage}

    \vspace{0.5em}

    \begin{minipage}[gt]{0.48\linewidth}
        \centering
        \includegraphics[width=\linewidth]{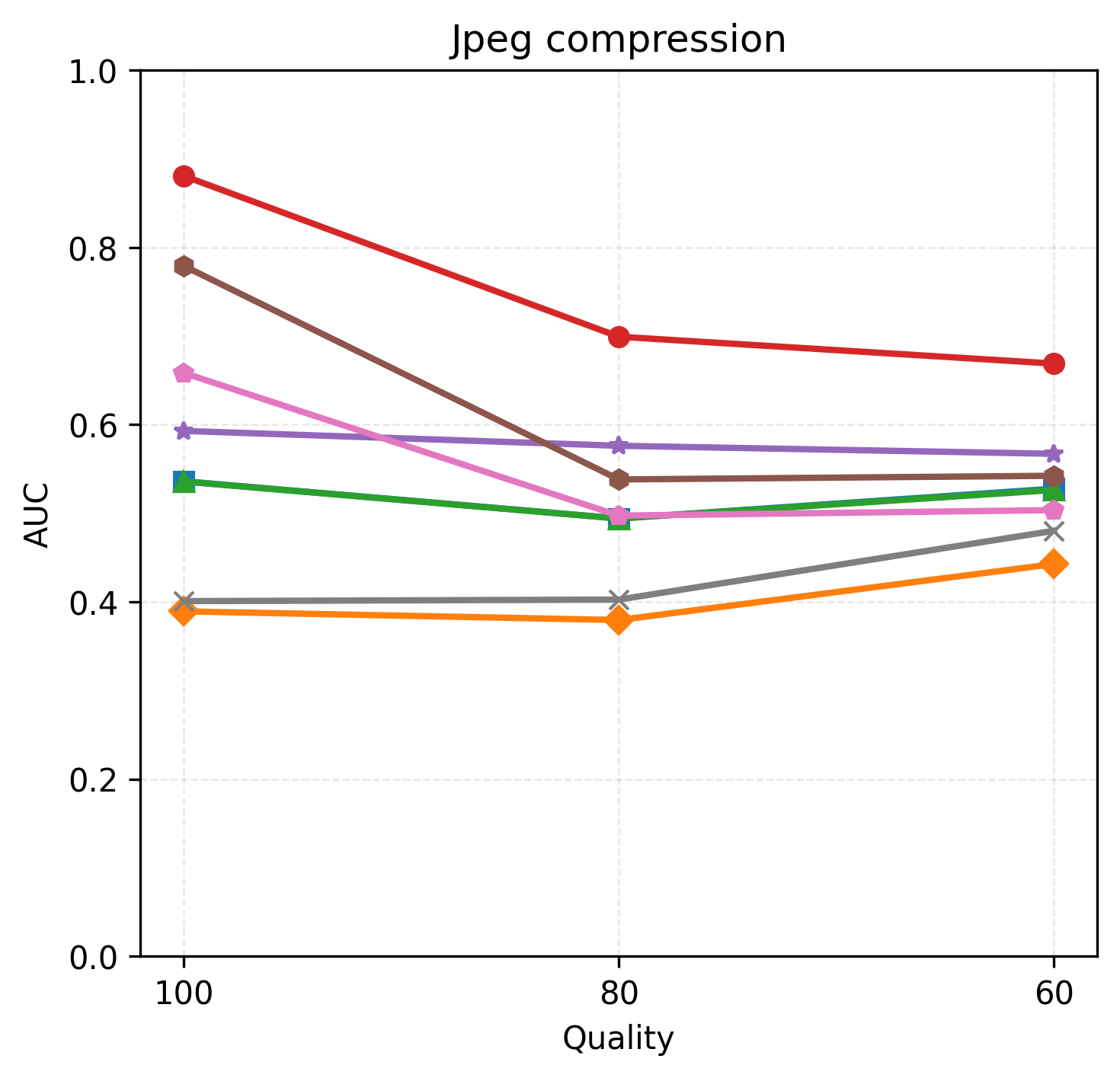}
        \subcaption{JPEG Compression}
    \end{minipage}
    \hfill
    \begin{minipage}[gt]{0.48\linewidth}
        \centering
        \includegraphics[width=\linewidth]{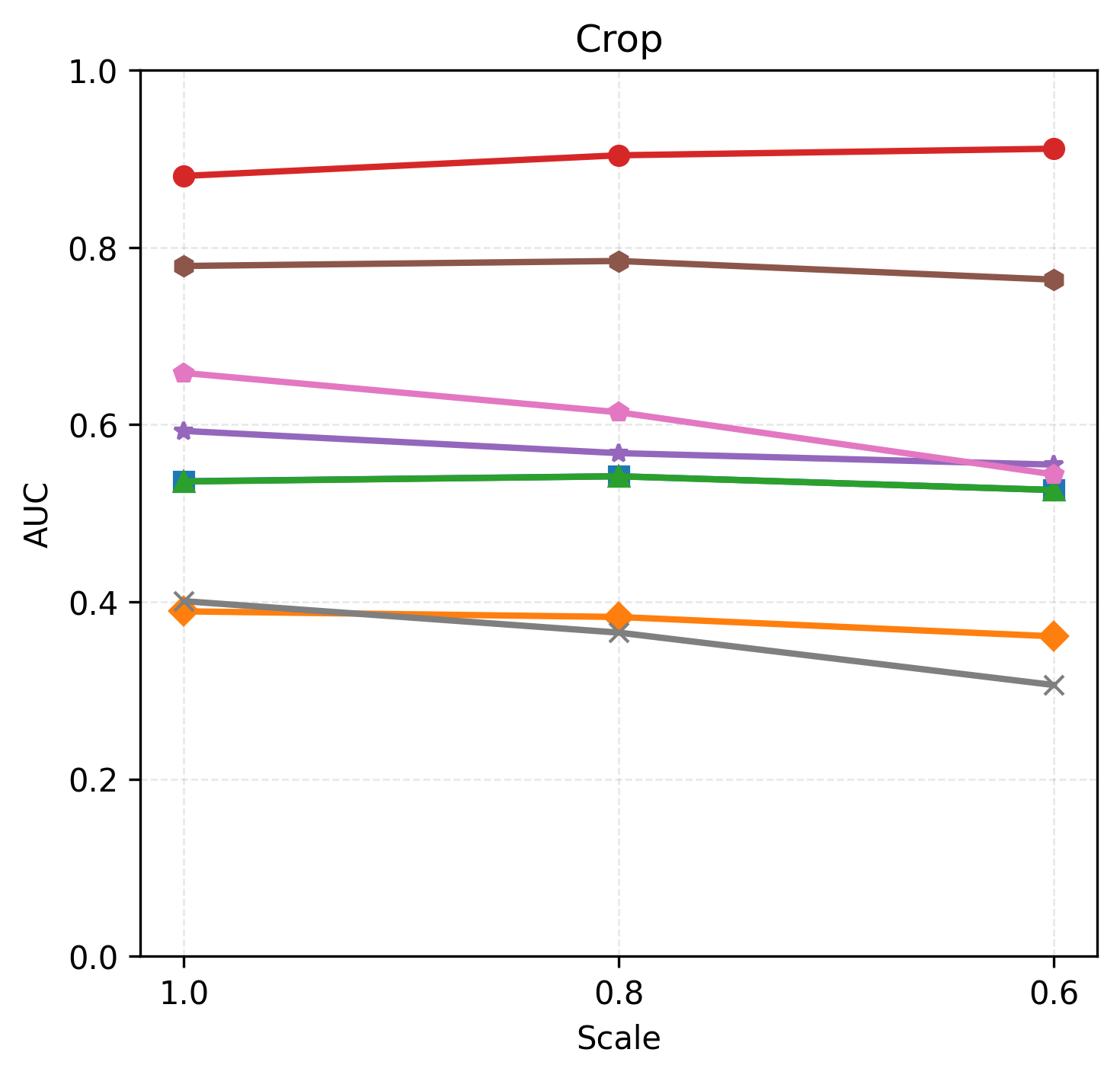}
        \subcaption{Crop}
    \end{minipage}

    \vspace{0.4em}
    \begin{minipage}{.9\linewidth}
        \centering
        \includegraphics[width=0.9\linewidth]{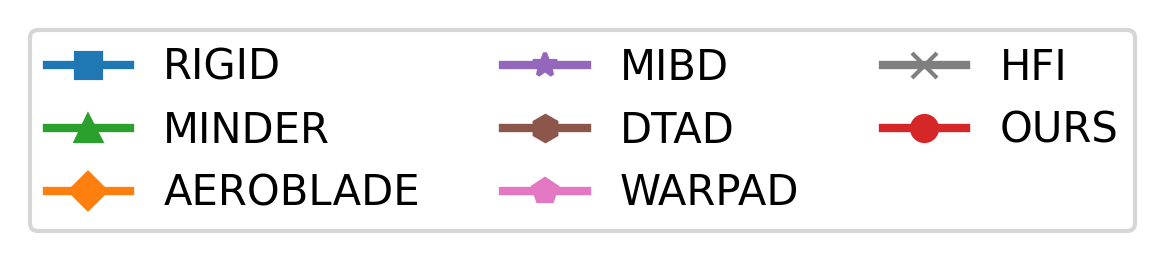}
    \end{minipage}

    \caption{Robustness evaluation across common perturbations, measured in AUC.}
    \label{fig:robustness_evaluation}
\end{figure}

\subsection{Ablation Studies}\label{subsec:ablation}
For a deeper analysis of the proposed method, we investigate the robustness to image corruption, the impact of hyperparameters, and the influence of different vision models.

\noindent\textbf{Robustness to Corruptions}

Robustness to post-processing artifacts is critical for synthetic image detection, 
as images shared on real-world platforms are frequently subject to compression, 
resizing, and other degradations that alter low-level statistics. 
To assess robustness under such distribution shifts, we evaluate performance 
under four common perturbations: Gaussian blur, Gaussian noise, center cropping, 
and JPEG compression.
We do not conduct ablation studies for ZEROFAKE~\cite{zerofake} due to its substantially higher computational cost (Table~\ref{tab:runtime}), which makes repeated evaluations impractical under our experimental setup.

Experiments are conducted on the OpenFake dataset, where each perturbation 
is applied at three increasing intensity levels.
Figure~\ref{fig:robustness_evaluation} reports the overall AUC for each perturbation.
Existing methods are often affected by perturbations, thus reducing their performances.
For example, DTAD exhibits substantial degradation under JPEG compression, with performance approaching random classifier at higher compression levels. 
Other representative methods show unstable behavior across perturbation types, with AUC frequently dropping below $0.5$, indicating limited robustness to distributional corruption.
Our method demonstrates stable performance across most perturbations.
Although Gaussian noise and JPEG compression leads to a noticeable decline in detection accuracy when the intensity increases, the overall degradation remains moderate.
Across all perturbation types and intensity levels, our method achieves the highest 
average AUC, indicating superior robustness under realistic image degradation scenarios.



\noindent\textbf{Hyperparameter Analysis}

We analyze the sensitivity of our method to its key hyperparameters, including noise strength $\lambda$, the patch-size $P$, and the selected CLIP layer $l$.

\begin{figure}[ht]
    \centering
    \begin{minipage}[t]{0.32\linewidth}
        \centering
        \includegraphics[width=\linewidth]{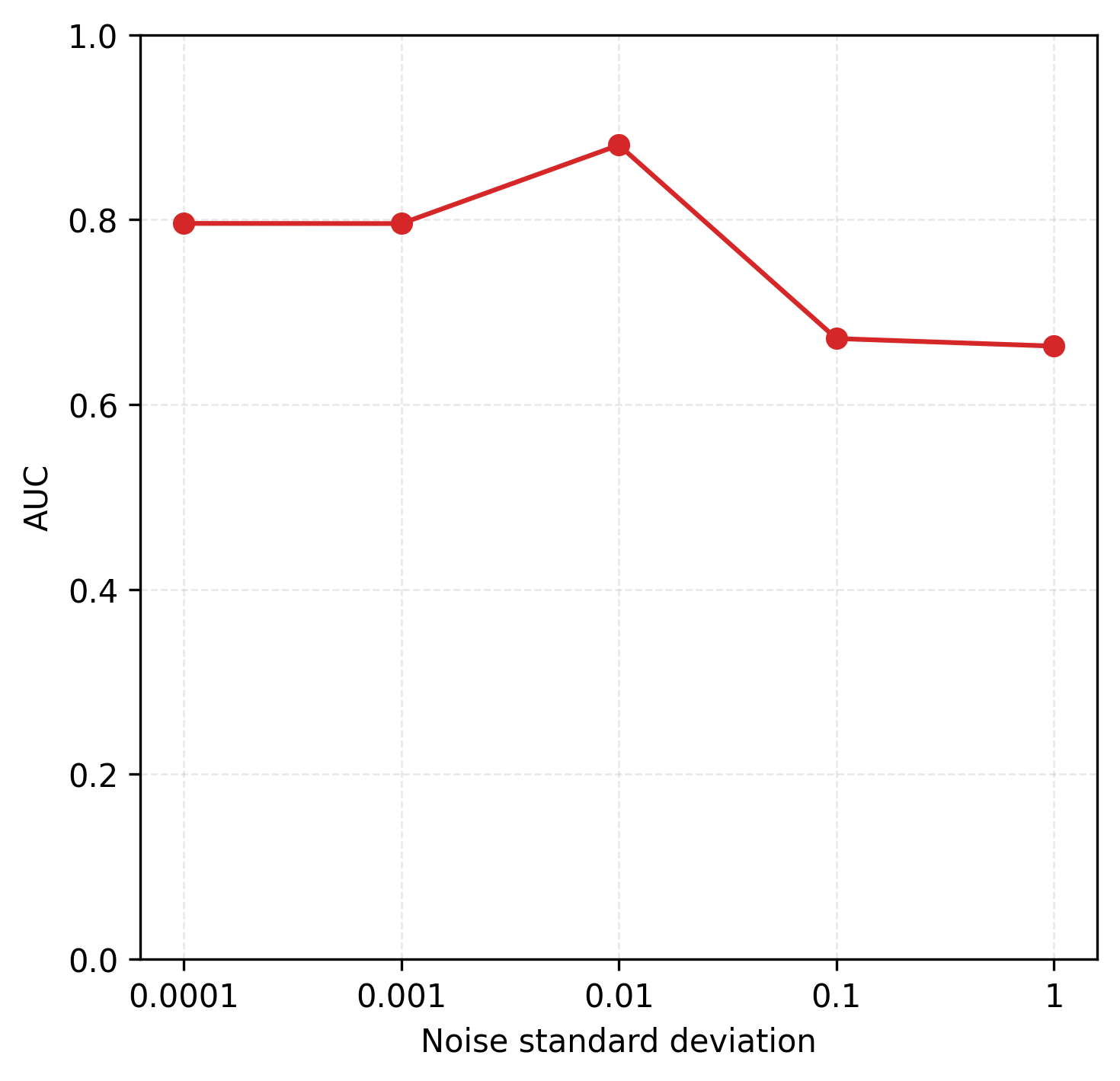} 
        \subcaption{Noise strength $\lambda$}
    \end{minipage}
    \hfill
    \begin{minipage}[t]{0.32\linewidth}
        \centering
        \includegraphics[width=\linewidth]{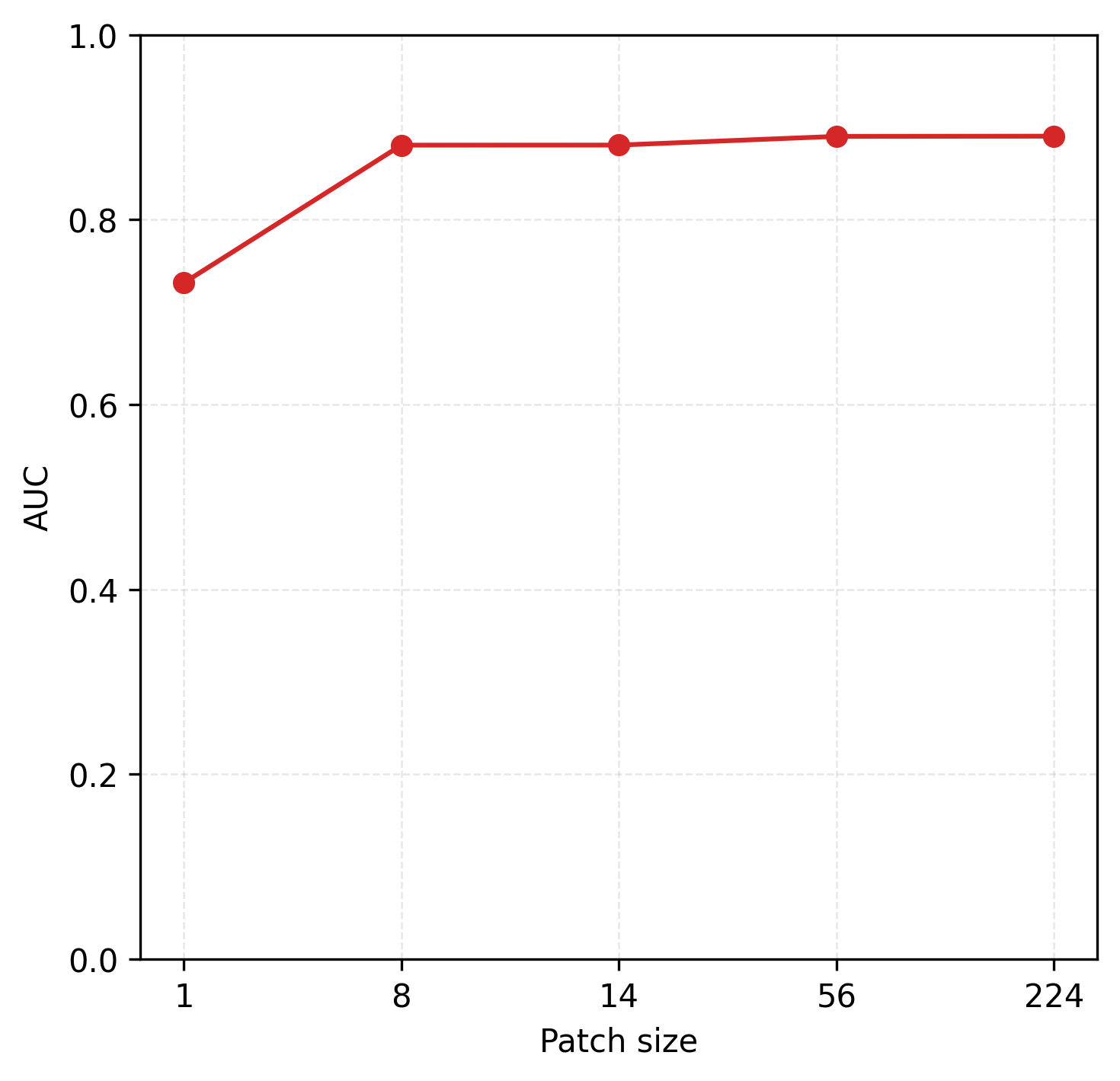} 
        \subcaption{Patch size $P$}
    \end{minipage}
    \hfill
    \begin{minipage}[t]{0.32\linewidth}
        \centering
        \includegraphics[width=\linewidth]{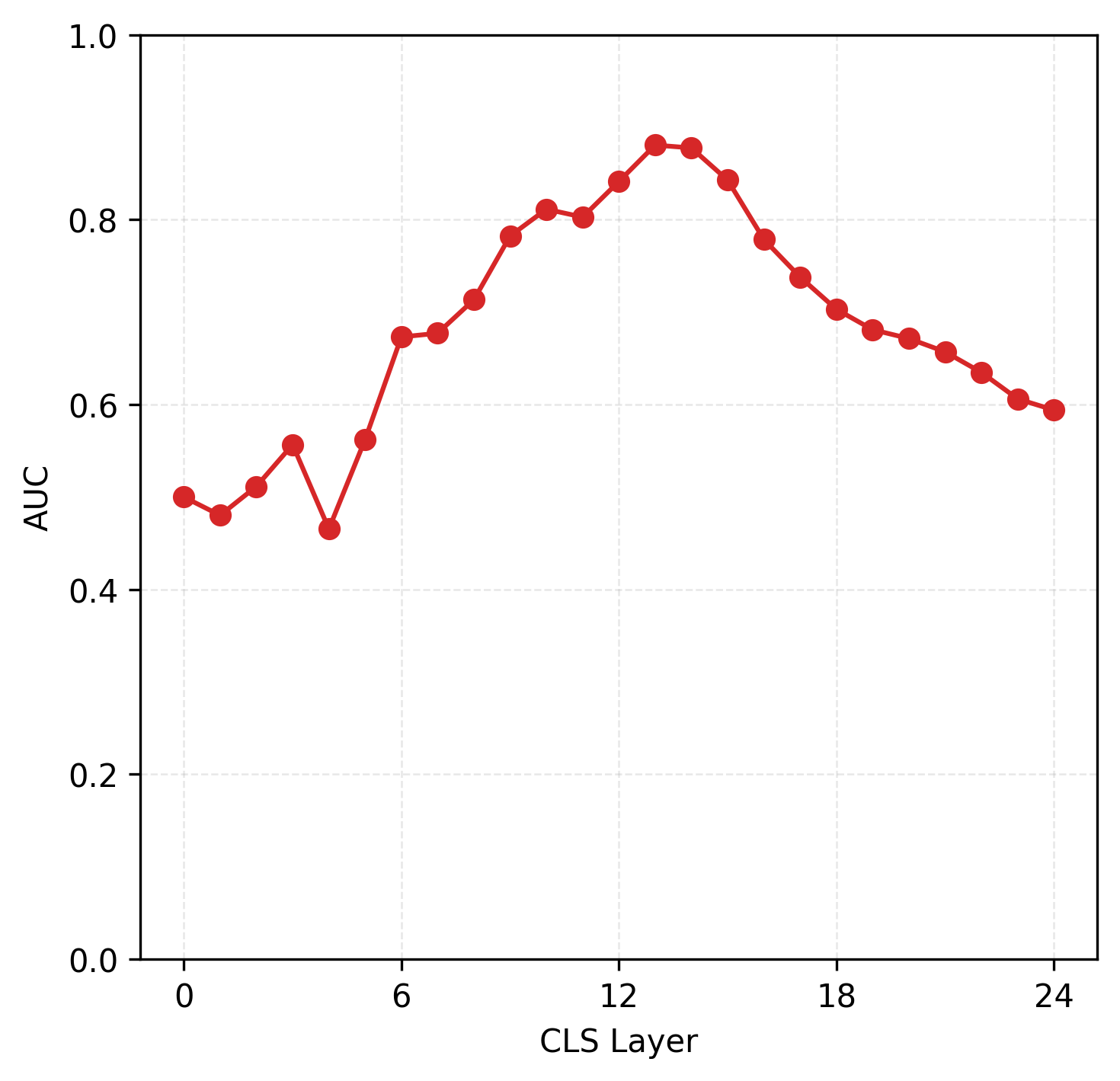} 
        \subcaption{CLS layer $l$}
    \end{minipage}
    \caption{Hyperparameter analysis of our method w.r.t. AUC result on Openfake dataset.}
    \label{fig:hyperparameter_analysis}
\end{figure}

Figure~\ref{fig:hyperparameter_analysis}a examines the effect of the noise magnitude $\lambda$ on the OpenFake dataset.
Excessive noise levels ($\lambda = 1.0$ and $0.1$) lead to a clear degradation in AUC, indicating that overly strong perturbations distort discriminative cues necessary for detection.
Performance peaks at $\lambda = 0.01$, with further reduction in noise strength resulting in only marginal decline.
This suggests the existence of an optimal perturbation scale that maximizes 
discriminative sensitivity, whereas excessive noise distorts image statistics and reduces discriminative cues.

Figure~\ref{fig:hyperparameter_analysis}b evaluates the impact of patch granularity.
We observe that detection performance varies with the spatial granularity of perturbation.
Extremely fine-grained (pixel-wise) perturbations degrade performance ($P=1$),
whereas intermediate-to-global perturbations yield stronger discriminative responses.
This suggests that synthetic artifacts manifest at intermediate-to-global frequency scales rather than at purely local structures.

We further investigate the influence of the selected CLIP ViT layer $l$.
Features extracted from different layers exhibit distinct representational properties.
Shallow layers primarily encode low-level texture statistics, while deeper layers capture high-level semantic abstractions~\cite{jiang2023clip}.
As shown in Figure~\ref{fig:hyperparameter_analysis}c, shallow layers yield AUC values close to random performance, suggesting that purely low-level cues are insufficient for reliable discrimination.
Conversely, the deepest layers also produce suboptimal performance, implying that high-level semantic representations alone do not capture the artifacts introduced by generative models.
The best performance is achieved around layer $13$, indicating that intermediate representations provide the most discriminative signal.
This observation suggests that synthetic artifacts are encoded in mid-level features,
which balance structural and frequency-sensitive information.

\FloatBarrier
\begin{table}[ht]
    \centering
    \caption{Proposed method's performance (AUC) and runtime in the Openfake dataset under different backbones. For each model, we extract the feature representations from the layer with the highest AUC.}
    \label{tab:model}
    \input{tables/model}
\end{table}
\noindent\textbf{Effect of models}

Table~\ref{tab:model} reports the impact of different vision foundation models on detection performance.

We first evaluate the smaller variant of CLIP model (ViT-B/32~\footnote{https://huggingface.co/openai/clip-vit-base-patch32}), which achieves improved inference speed compared to ViT-L/14 but its detection accuracy on OpenFake is lower.
This suggests that our method benefits from higher-capacity representations, which better capture the subtle frequency discrepancies exploited by our perturbation strategy.
We further investigate self-supervised vision foundation models, including DINOv2~\footnote{https://huggingface.co/facebook/dinov2-large} and DINOv3~\footnote{https://huggingface.co/facebook/dinov3-vits16-pretrain-lvd1689m}.
Both models yield substantially lower performance than CLIP-based variants.
A plausible explanation is that their training paradigm emphasizes robustness and invariance to local perturbations, which may suppress sensitivity to the fine-grained frequency artifacts essential for distinguishing synthetic images from natural ones.
We also observe a consistent runtime difference between CLIP models and DINO models under the same experimental setting.







%% file: tables/openfake.tex
\begin{tabular}{lccccccccc}\toprule
Generator & RIGID & MINDER & \makecell[l]{AERO\\BLADE} & MIBD & \makecell[l]{ZERO\\FAKE} & DTAD & \makecell[l]{WAR\\PAD} & HFI & OURS \\
\midrule
Aurora-20-1-25 & 0.473 & 0.471 & 0.343 & 0.788 & 0.684 & 0.228 & 0.420 & 0.487 & \textbf{0.850} \\
Chroma & 0.633 & 0.633 & 0.452 & 0.503 & 0.571 & \textbf{0.924} & 0.728 & 0.432 & 0.879 \\
DALLE-3 & 0.469 & 0.468 & 0.311 & 0.572 & 0.560 & 0.386 & 0.525 & 0.354 & \textbf{0.858} \\
Flux-1.1-pro & 0.434 & 0.434 & 0.531 & 0.445 & 0.605 & 0.485 & 0.567 & 0.527 & \textbf{0.808} \\
\makecell[l]{Flux-amateur\\snapshotphotos} & 0.506 & 0.505 & 0.559 & 0.460 & 0.502 & \textbf{0.934} & 0.627 & 0.498 & 0.797 \\
Flux-mvc5000 & 0.371 & 0.371 & 0.560 & 0.501 & 0.543 & 0.772 & 0.381 & 0.476 & \textbf{0.857} \\
Flux-realism & 0.548 & 0.547 & 0.430 & 0.488 & 0.719 & \textbf{0.909} & 0.608 & 0.415 & 0.814 \\
Flux.1-dev & 0.558 & 0.557 & 0.489 & 0.481 & 0.694 & \textbf{0.942} & 0.627 & 0.522 & 0.790 \\
Flux.1-schnell & 0.577 & 0.575 & 0.591 & 0.485 & 0.633 & \textbf{0.948} & 0.724 & 0.595 & 0.836 \\
Frames-23-1-25 & 0.440 & 0.438 & 0.340 & 0.832 & 0.550 & \textbf{0.900} & 0.660 & 0.376 & 0.860 \\
GPT-image-1 & 0.512 & 0.510 & 0.358 & 0.548 & 0.675 & 0.700 & 0.543 & 0.400 & \textbf{0.857} \\
Grok-2-image-1212 & 0.503 & 0.501 & 0.446 & 0.474 & 0.694 & 0.396 & 0.298 & 0.516 & \textbf{0.700} \\
Halfmoon-4-4-25 & 0.546 & 0.544 & 0.422 & 0.855 & 0.438 & \textbf{0.868} & 0.750 & 0.459 & 0.847 \\
Hidream-i1-full & 0.552 & 0.554 & 0.521 & 0.539 & 0.541 & \textbf{0.918} & 0.690 & 0.487 & 0.866 \\
Ideogram-2.0 & 0.486 & 0.484 & 0.390 & 0.809 & 0.485 & 0.852 & 0.713 & 0.382 & \textbf{0.901} \\
Ideogram-3.0 & 0.591 & 0.589 & 0.389 & 0.478 & 0.391 & \textbf{0.932} & 0.685 & 0.489 & 0.897 \\
Imagen-3.0-002 & 0.442 & 0.440 & 0.313 & 0.462 & 0.490 & 0.845 & 0.517 & 0.379 & \textbf{0.930} \\
Imagen-4.0 & 0.467 & 0.465 & 0.410 & 0.419 & 0.501 & 0.790 & 0.572 & 0.384 & \textbf{0.918} \\
Lumina-17-2-25 & 0.620 & 0.623 & 0.591 & 0.844 & 0.683 & \textbf{0.928} & 0.744 & 0.570 & 0.897 \\
Midjourney-6 & 0.435 & 0.433 & 0.260 & \textbf{0.792} & 0.447 & 0.315 & 0.504 & 0.327 & 0.766 \\
Midjourney-7 & 0.636 & 0.635 & 0.236 & 0.740 & 0.627 & 0.872 & 0.723 & 0.449 & \textbf{0.877} \\
Mystic & 0.645 & 0.644 & 0.468 & 0.486 & 0.645 & 0.908 & 0.687 & 0.326 & \textbf{0.915} \\
Recraft-v2 & 0.322 & 0.319 & 0.396 & 0.748 & 0.504 & 0.707 & 0.668 & 0.328 & \textbf{0.960} \\
Recraft-v3 & 0.394 & 0.391 & 0.308 & 0.821 & 0.730 & 0.623 & 0.764 & 0.282 & \textbf{0.958} \\
SD1.5 & 0.524 & 0.537 & 0.267 & 0.847 & 0.297 & 0.684 & 0.792 & 0.336 & \textbf{0.952} \\
\makecell[l]{SD1.5-dream\\shaper} & 0.721 & 0.720 & 0.190 & 0.663 & 0.468 & 0.764 & 0.745 & 0.259 & \textbf{0.885} \\
SD1.5-epicdream & 0.523 & 0.521 & 0.257 & 0.594 & 0.708 & 0.758 & 0.723 & 0.260 & \textbf{0.907} \\
SD2.1 & 0.724 & 0.728 & 0.375 & 0.865 & 0.522 & 0.871 & 0.842 & 0.418 & \textbf{0.947} \\
SD3.5 & 0.490 & 0.488 & 0.511 & 0.568 & 0.471 & \textbf{0.956} & 0.733 & 0.375 & 0.904 \\
SDXL & 0.695 & 0.694 & 0.410 & 0.646 & 0.681 & 0.895 & 0.838 & 0.355 & \textbf{0.924} \\
SDXL-epic-realism & 0.516 & 0.515 & 0.202 & 0.579 & 0.673 & 0.800 & 0.689 & 0.289 & \textbf{0.973} \\
SDXL-juggernaut & 0.492 & 0.492 & 0.367 & 0.521 & 0.633 & 0.908 & 0.683 & 0.380 & \textbf{0.951} \\
SDXL-realvis-v5 & 0.434 & 0.433 & 0.326 & 0.607 & 0.431 & 0.754 & 0.803 & 0.301 & \textbf{0.923} \\
\makecell[l]{SDXL-touchof\\realism} & 0.689 & 0.688 & 0.272 & 0.610 & 0.781 & 0.853 & 0.802 & 0.298 & \textbf{0.931} \\
\hline
Average & 0.536 & 0.536 & 0.389 & 0.593 & 0.580 & 0.779 & 0.658 & 0.401 & \textbf{0.881} \\
\bottomrule
\end{tabular}

%% file: tables/semitruth.tex

\begin{tabular}{lccccccccc}\toprule
Generator & RIGID & MINDER & \makecell[l]{AERO\\BLADE} & MIBD & \makecell[l]{ZERO\\FAKE} & DTAD & \makecell[l]{WAR\\PAD} & HFI & OURS \\
\midrule
Kandinsky22 & 0.450 & 0.451 & 0.723 & 0.548 & 0.507 & 0.829 & \textbf{0.852} & 0.606 & 0.844 \\
OpenJourney & 0.498 & 0.498 & 0.707 & 0.499 & 0.562 & 0.649 & 0.680 & 0.700 & \textbf{0.801} \\
SDXL & 0.352 & 0.352 & 0.531 & 0.482 & 0.525 & 0.686 & 0.805 & 0.659 & \textbf{0.834} \\
SD1.4 & 0.350 & 0.351 & 0.686 & 0.525 & 0.679 & 0.730 & 0.750 & \textbf{0.975} & 0.872 \\
SD1.5 & 0.447 & 0.448 & 0.724 & 0.530 & 0.668 & 0.780 & 0.785 & \textbf{0.967} & 0.854 \\
\hline
Average & 0.418 & 0.419 & 0.671 & 0.517 & 0.583 & 0.736 & 0.779 & 0.771 & \textbf{0.841} \\
\bottomrule
\end{tabular}

%% file: tables/genimage.tex

\begin{tabular}{lccccccccc}\toprule
Generator & RIGID & MINDER & \makecell[l]{AERO\\BLADE} & MIBD & \makecell[l]{ZERO\\FAKE} & DTAD & \makecell[l]{WAR\\PAD} & HFI & OURS \\
\midrule
ADM & 0.845 & 0.850 & 0.690 & 0.754 & 0.918 & 0.935 & \textbf{0.989} & 0.759 & 0.974 \\
BigGAN & 0.961 & 0.961 & 0.839 & 0.942 & 0.901 & 0.973 & \textbf{0.999} & 0.950 & 0.886 \\
Glide & 0.901 & 0.900 & 0.760 & 0.876 & 0.924 & 0.968 & \textbf{0.992} & 0.863 & 0.979 \\
Midjourney & 0.684 & 0.681 & 0.595 & 0.565 & 0.781 & 0.923 & 0.840 & 0.576 & \textbf{0.972} \\
SD1.4 & 0.603 & 0.606 & 0.835 & 0.717 & 0.904 & 0.950 & 0.955 & 0.937 & \textbf{0.991} \\
SD1.5 & 0.596 & 0.600 & 0.833 & 0.720 & 0.907 & 0.949 & 0.950 & 0.943 & \textbf{0.991} \\
VQDM & 0.911 & 0.918 & 0.600 & 0.901 & 0.794 & 0.918 & \textbf{0.986} & 0.693 & 0.985 \\
Wukong & 0.630 & 0.643 & 0.870 & 0.704 & 0.837 & 0.935 & 0.941 & 0.959 & \textbf{0.992} \\
\hline
Average & 0.760 & 0.763 & 0.756 & 0.770 & 0.873 & 0.944 & 0.956 & 0.839 & \textbf{0.972} \\
\bottomrule
\end{tabular}

%% file: tables/runtime.tex


\begin{tabular}{lccc}\toprule
Method & Openfake & Semi-Truth & Genimage \\
\midrule
RIGID & 1519.485 & 860.492 & 2550.385 \\
MINDER & 3019.285 & 1707.962 & 5055.123 \\
AEROBLADE & 11042.416 & 6246.216 & 18476.045 \\
MIBD & 75610.938 & 42668.291 & 126113.634 \\
ZEROFAKE & 2095768.396 & 1184961.851 & 3510376.209 \\
DTAD & 276861.686 & 156680.426 & 463905.786 \\
WARPAD & 19933.757 & 11283.683 & 33435.179 \\
HFI & 6568.457 & 3710.264 & 11053.292 \\
OURS & 436.752 & 190.816 & 559.813 \\
\bottomrule
\end{tabular}

%% file: tables/model.tex
\begin{tabular}{lccc}\hline
        Model & AUC & Runtime \\ \hline
         CLIP-B/32  (Layer 8) & 0.7652 & 245.3 \\ 
         DINOv2 (Layer 14) & 0.6795 & 1437.6 \\ 
         DINOv3 (Layer 6) & 0.5019 & 1172.4 \\\hline 
         CLIP-L/14 (Layer 13) & 0.881 & 436.8 \\ \hline
\end{tabular}

%% file: sections/conclusion.tex
\section{Conclusion}
We presented a training-free method for AI-generated image detection that exploits structured high-frequency perturbations and semantic representations from vision-language foundation models. 
By measuring representation sensitivity at an intermediate layer, our approach effectively captures fine-grained discrepancies between real and generated images that are often overlooked in existing training-free detectors. 
Extensive experiments demonstrate that our method achieves improved detection accuracy and robustness across diverse benchmarks and generative models, while requiring no additional training or model adaptation. 
Owing to its lightweight perturbation design, the proposed approach is computationally efficient and readily scalable to real-world deployment.

\noindent\textbf{Limitations and Future Work}

Our current evaluation focuses on ranking performance. 
Future work will investigate how to determine an appropriate decision threshold for practical deployment settings.
Moreover, the current framework relies on a fixed frequency decomposition; exploring adaptive frequency representations may further improve robustness. 
Finally, extending the approach to incorporate cross-modal consistency signals, such as text-image alignment, could enhance resilience against increasingly sophisticated generative models.